\documentclass{article}%
\usepackage{amsmath}
\usepackage{amsfonts}
\usepackage{amssymb}
\usepackage{graphicx}
\usepackage{algorithmic}
\usepackage{algorithm}
\usepackage{hyperref}
\providecommand{\U}[1]{\protect \rule{.1in}{.1in}}
\providecommand{\U}[1]{\protect \rule{.1in}{.1in}}

\begin{document}

\title{SurvBETA: Ensemble-Based Survival Models Using Beran Estimators and Several
Attention Mechanisms}
\author{Lev V. Utkin, Semen P. Khomets, Vlada A. Efremenko and Andrei V. Konstantinov\\Higher School of Artificial Intelligence Technologies\\Peter the Great St.Petersburg Polytechnic University\\St.Petersburg, Russia\\e-mail: lev.utkin@gmail.com, homets\_sp@spbstu.ru, \\
efremenko\_va@spbstu.ru, andrue.konst@gmail.com}
\date{}
\maketitle

\begin{abstract}
Many ensemble-based models have been proposed to solve machine learning
problems in the survival analysis framework, including random survival
forests, the gradient boosting machine with weak survival models, ensembles of
the Cox models. To extend the set of models, a new ensemble-based model called
SurvBETA (the Survival Beran estimator Ensemble using Three Attention mechanisms)
is proposed where the Beran estimator is used as a weak learner in the
ensemble. The Beran estimator can be regarded as a kernel regression model
taking into account the relationship between instances. Outputs of weak
learners in the form of conditional survival functions are aggregated with
attention weights taking into account the distance between the analyzed
instance and prototypes of all bootstrap samples. The attention mechanism is
used three times: for implementation of the Beran estimators, for determining
specific prototypes of bootstrap samples and for aggregating the weak model
predictions. The proposed model is presented in two forms: in a general form
requiring to solve a complex optimization problem for its training; in a
simplified form by considering a special representation of the attention
weights by means of the imprecise Huber's contamination model which leads to
solving a simple optimization problem. Numerical experiments illustrate
properties of the model on synthetic data and compare the model with other
survival models on real data. A code implementing the proposed model is
publicly available.

\end{abstract}

\section{Introduction}

In spite of developing many efficient machine learning models, the
ensemble-based models can be regarded as one of the most powerful tools for
solving various tasks. The main idea behind the ensemble-based models is based
on combining a set of weak or base models to construct a strong model having a
more accurate and generalizable prediction. The most famous ensemble-based
models are Random Forests \cite{Breiman-2001} and their numerous modifications
based on the bagging technique \cite{Breiman-1996}, AdaBoost
\cite{Freund-Shapire-97} and the Gradient Boosting Machines
\cite{Friedman-2001,Friedman-2002}, which use the boosting technique. The
above ensemble-based models mainly differ in weak models or in various
approaches to combining the weak model predictions. It is interesting to point
out that many modern models contain some elements of the ensemble-based
methodology, for example, the well-known multi-head attention
\cite{Vaswani-etal-17} used in transformers can also be viewed as an ensemble
of the attention mechanisms.

The efficiency and simplicity of ensemble-based models motivated to develop a
huge amount of new models. Surveys and detailed descriptions of the
ensemble-based methodology and various models can be found in
\cite{Xibin_Dong-etal-20,Ferreira-Figueiredo-2012,Ren-Zhang-Suganthan-2016,Sagi-Rokach-2018,Wozniak-etal-2014,ZH-Zhou-2012}%
.

An important area of the ensemble-based model application and development is
survival analysis which deals with a specific type of data, namely, with
censored survival data. Survival analysis studies observations in the form of
times to an event of interest \cite{Wang-Li-Reddy-2019}. Moreover, events
might not have happened during the observation period \cite{Nezhad-etal-2018}.
In this case, the corresponding observed times is less than or equal to a true
event time, and the observation is called censored. This implies that every
dataset in survival analysis consists of two types of objects depending on the
possibility to observe the event of interest: censored and uncensored.

Many ensemble-based models have been proposed to deal with survival data. A
large part of the models can be regarded as survival extensions of the
well-known ensemble-based models. In particular, random survival forests
(RSFs)
\cite{Ibrahim-etal-2008,Mogensen-etal-2012,Schmid-etal-2016,Wang-Zhou-2017,Utkin-etal-2019c,Utkin-Konstantinov-22d,Wright-etal-2017}
are extensions of the random forests to survival data. Survival models
proposed in \cite{chen2013gradient,liu2019hitboost} are extensions of the
gradient boosting machine.

An interesting ensemble-based model using the Cox proportional hazard model
\cite{Cox-1972} as a weak model was proposed by Meier et al.
\cite{Meier-etal-16}. The Cox model and its modifications are successfully
applied to many survival problems. A reason of the success is that the Cox
model takes into account feature vectors and allows us to obtain conditional
survival measures. However, the Cox model does not take into account relative
positions of feature vectors. For example, the Cox model may provide incorrect
survival measures when a dataset has a multi-cluster structure. In order to
overcome this limitation and to construct a more adequate model, we propose to
apply the Beran estimator \cite{Beran-81} which estimates the conditional
survival function taking into account the relationship between feature vectors
by assigning weights to instances depending on their proximity to the analyzed
instance. It is implemented by means of kernels. Therefore, the Beran
estimator can be viewed a type of the kernel regression.

The above positive property of the Beran estimator motivates to develop new
ensemble-based models where the Beran estimators as weak learners. A simple
way to implement an ensemble of the Beran estimators is to use the standard
bagging approach and to average obtained predictions in the form of survival
functions provided by each Beran estimator from the ensemble. However,
survival functions predicted by different weak models may be sufficiently
different especially when the weak models are trained on instances from
different clusters. Therefore, in order to improve the whole ensemble-based
model, we propose to apply the attention mechanism to predictions in the form
of SFs such that the attention weights are determined through the feature
vectors. The main idea to aggregate the conditional SFs is to take into
account the distance between the analyzed feature vector and prototypes of all
bootstrap samples obtained by using the attention mechanism in the form of the
Nadaraya-Watson regression \cite{Nadaraya-1964,Watson-1964}. It is important to point out that prototypes of each bootstrap subsample are determined depending on the analyzed feature vector by means of the attention mechanism, i.e., each bootstrap subsample can have different prototypes for different testing instances.
As a result, we apply the attention mechanism to the ensemble-based model three times: for implementation of the Beran estimators, for determining specific prototypes of bootstrap samples and for aggregating the weak model
predictions.

Another important difference between the proposed model and other ensemble-based models is its specific bootstrap sampling scheme. It is implemented in such a way that subsamples consist of instances that are close to each other and, at the same time, have certain areas of intersection with some of other subsamples, which is required for the correct use of prototypes.

The proposed ensemble-based survival model is called SurvBETA (the
\textbf{Surv}ival \textbf{B}eran estimator \textbf{E}nsemble using 
\textbf{T}hree \textbf{A}ttention mechanisms). Our contributions can be summarized as follows:

\begin{enumerate}
\item A new bagging-based survival model using the Beran estimator as a weak
model is proposed for predicting survival measures. The main idea behind the
model is to aggregate the weak model predictions applying the attention mechanisms.

\item The proposed model is significantly simplified by considering a special
representation of the attention weights by means of the imprecise Huber's
$\epsilon$-contamination model \cite{Huber81}.

\item A specific scheme for generating the bootstrap subsamples is proposed
that ensures comparable contributions of different Beran estimators from the ensemble,

\item Various numerical experiments compare the method with other survival
models, including the RSF, GBM Cox \cite{ridgeway1999state}, GBM AFT
\cite{Barnwal-etal-22}, and the single Beran estimator.
\end{enumerate}

A code implementing the proposed model is publicly available at:
\url{https://github.com/NTAILab/SurvBETA}.

\section{Related work}

\textbf{Machine learning models in survival analysis}. Detailed reviews of
many survival models can be found in \cite{Wang-Li-Reddy-2019,Wiegrebe:2024aa}%
. Most models can be viewed as extensions of the base machine learning models
when survival data is available. For example, the RSFs are based on ideas of
the random forest, but it uses the specific node splitting rules and the
specific calculation of predictions. Many survival models modify the Cox model
by replacing the linear relationship of covariate with other machine learning
models, for example, \cite{Tibshirani-1997} replaced this relationship with
the Lasso method. Similar replacement were proposed in order to take into
account a high dimensionality of survival data. Following this paper, several
modifications of the Lasso methods for censored data were proposed
\cite{Kaneko-etal-2015,Ternes-etal-2016}. Many survival models are based on
applying neural networks and deep learning
\cite{Katzman-etal-2018,Luck-etal-2017,Nezhad-etal-2018,Yao-Zhu-Zhu-Huang-2017}%
. Deep recurrent survival model was proposed in \cite{ren2019deep}. An
interesting approach to add the pairwise ranking loss function based on
ranking information of survival data to training of the deep survival model
was considered in \cite{jing2019deep}. The convolutional neural networks also
have been applied to the survival analysis \cite{Haarburger-etal-2018}.
Transformer-based survival models were studied in
\cite{hu2021transformer,tang2023explainable,Wang-Sun-22}.

It should be noted that the survival model based on neural networks require a
large amount of training data. This can be a problem in many applications.
Moreover, tabular data can also encounter difficulties when survival neural
networks are used to predict. Therefore, ensemble-based survival models using
simple weak leaners may provide better results for several cases.

\textbf{Ensemble-based survival machine learning models.} Starting from works
\cite{Ishwaran-etal-2008} and \cite{Ishwaran-Kogalur-2007}, many approaches to
constructing ensembles of survival models have been proposed. A fundamental
work stating basic principles of survival ensembles was presented by Hothorn
et al. \cite{Hothorn-etal-2006}. Salerno et al. \cite{salerno2023high} reviewed
various survival models dealing with high-dimensional data, including
ensemble-based survival models. Ensembles of survival models with
high-dimensional data were also studied in \cite{wang2017selective}.
Time-varying covariates treated by ensemble-based survival models were studied
in \cite{yao2022ensemble}. Bayesian survival tree ensembles were considered in
\cite{linero2022bayesian}. Various ensemble-based survival models were also
presented in
\cite{chen2013gradient,liu2019hitboost,Meier-etal-16,Mogensen-etal-2012,Schmid-etal-2016,Wang-Zhou-2017,Utkin-etal-2019c,Utkin-Konstantinov-22d,Wright-etal-2017}%
. However, it is interesting to point out that there are no ensemble-based
models which use the Beran estimator as a weak model. At the same time, our
experiments show that models based on the Beran estimators can compete with
other models in case of complex data structure.

\section{Elements of survival analysis}

\subsection{Basic concepts}

An instance (object) in survival analysis is often represented by a triplet
$(\mathbf{x}_{i},\delta_{i},T_{i})$, where $\mathbf{x}_{i}^{\mathrm{T}%
}=(x_{i1},...,x_{id})\in \mathbb{R}^{d}$ is the vector of the instance
features; $T_{i}\in \mathbb{R}$ is the event time for the $i$-th instance. In
survival analysis, all events can be divided into two groups
\cite{Hosmer-Lemeshow-May-2008}. The first group contains observed events
called uncensored. For these events, $T_{i}$ is the time between a baseline
time and the time of event happening. The corresponding instances are
identified by the censoring indicator $\delta_{i}=1$. The second group
consists of unobserved events. In this case, $T_{i}$ is the time between the
baseline time and the end of the observation. The corresponding instances are
identified by the censoring indicator $\delta_{i}=0$.

Let a training set $\mathcal{A}$ consist of $n$ triplets $(\mathbf{x}%
_{i},\delta_{i},T_{i})$, $i=1,...,n$. Survival analysis aims to estimate the
event time $T$ for a new instance $\mathbf{x}$ on the basis of the dataset
$\mathcal{A}$. Important concepts in survival analysis are survival functions
(SFs) and cumulative hazard functions (CHFs). The conditional SF denoted as
$S(t\mid \mathbf{x})$ is the probability of surviving up to time $t$, that is
$S(t\mid \mathbf{x})=\Pr \{T>t\mid \mathbf{x}\}$. The CHF denoted as
$H(t\mid \mathbf{x})$ is expressed through the SF as follows:
\begin{equation}
H(t\mid \mathbf{x})=-\ln S(t\mid \mathbf{x}).
\end{equation}

Comparison of survival models is often carried out by means of the C-index
\cite{Harrell-etal-1982}. It estimates the probability that the event times of
a pair of instances are correctly ranking. Let $\mathcal{J}$ is a set of all
pairs $(i,j)$ satisfying conditions $\delta_{i}=1$ and $T_{i}<T_{j}$. The
C-index is formally computed as \cite{Uno-etal-11,Wang-Li-Reddy-2019}:%
\begin{equation}
C=\frac{\sum_{(i,j)\in \mathcal{J}}\mathbf{1}[\widehat{T}_{i}<\widehat{T}_{j}%
]}{N}, \label{Survival_DF_24}%
\end{equation}
where $N$ is the number of all pairs $(i,j)$ from $\mathcal{J}$, i.e.,
$N=\sum_{(i,j)\in \mathcal{J}}1$; $\widehat{T}_{i}$ and $\widehat{T}_{j}$ are
expected event times.

\subsection{The Beran estimator}

According to the Beran estimator \cite{Beran-81}, the SF can be estimated on
the bases of the dataset $\mathcal{A}$ as follows:
\begin{equation}
S(t\mid \mathbf{x},\mathcal{A})=\prod_{t_{i}\leq t}\left \{  1-\frac
{\alpha(\mathbf{x},\mathbf{x}_{i})}{1-\sum_{j=1}^{i-1}\alpha(\mathbf{x}%
,\mathbf{x}_{j})}\right \}  ^{\delta_{i}}, \label{Beran_est}%
\end{equation}
where $t_{1}<t_{2}<...<t_{n}$ are ordered time moments obtained from the set
$\{T_{1},...,T_{n}\}$; the weight $\alpha(\mathbf{x},\mathbf{x}_{i})$
corresponds to location of $\mathbf{x}_{i}$ relative to $\mathbf{x}$ under
condition that the closer $\mathbf{x}_{i}$ to $\mathbf{x}$, the greater the
weight $\alpha(\mathbf{x},\mathbf{x}_{i})$.

The weights $\alpha(\mathbf{x},\mathbf{x}_{i})$ can be viewed as attention
weights and can be defined through kernels $K(\mathbf{x},\mathbf{x}_{i})$ as:%
\begin{equation}
\alpha(\mathbf{x},\mathbf{x}_{i})=\frac{K(\mathbf{x},\mathbf{x}_{i})}%
{\sum_{j=1}^{n}K(\mathbf{x},\mathbf{x}_{j})}.
\end{equation}

In particular, the Gaussian kernel with the parameter $\tau$ leads to the
weight:
\begin{equation}
\alpha(\mathbf{x},\mathbf{x}_{i})=\text{\textrm{softmax}}\left(
-\frac{\left \Vert \mathbf{x}-\mathbf{x}_{i}\right \Vert ^{2}}{\tau}\right)  .
\end{equation}

We can also write the CHF obtained by using the Beran estimator as:
\begin{equation}
H(t\mid \mathbf{x},\mathcal{A})=-\sum_{t_{i}\leq t}\delta_{i}\ln \left \{
\frac{1-\sum_{j=1}^{i}\alpha(\mathbf{x},\mathbf{x}_{j})}{1-\sum_{j=1}%
^{i-1}\alpha(\mathbf{x},\mathbf{x}_{j})}\right \}  ,
\end{equation}

The Beran estimator generalizes the Kaplan-Meier estimator
\cite{Wang-Li-Reddy-2019} for which $\alpha(\mathbf{x},\mathbf{x}_{i})=1/n$
for all $i=1,...,n$.

\section{A simple ensemble of Beran estimators}

Let $\mathcal{A}_{k}\subset \mathcal{A}$ be a random subset of the training set
$\mathcal{A}$ consisting of $m$ instances having indices from the set
$\mathcal{I}_{k}$. Let $t_{1}^{(k)}<...<t_{m}^{(k)}$ be the distinct times to
event of interest from the set $\{T_{j}:j\in \mathcal{I}_{k}\}$, where
$t_{1}^{(k)}=\min_{j\in \mathcal{I}_{k}}T_{j}$ and $t_{m}^{(k)}=\max
_{j\in \mathcal{I}_{k}}T_{j}$. It is important to note that the aggregation
rule in many implementations of random survival forests is based on averaging
the CHFs predicted by survival trees \cite{Ishwaran-etal-2008}. We also use
another approach by averaging SFs $S(t\mid \mathbf{x},\mathcal{A}_{k}%
)=S^{(k)}(t\mid \mathbf{x})$. SF $S^{(k)}(t\mid \mathbf{x})$ predicted by the
Beran estimator trained on the subset $\mathcal{A}_{k}$ is determined as:
\begin{align}
S(t  &  \mid \mathbf{x},\mathcal{A}_{k})=S^{(k)}(t\mid \mathbf{x})\nonumber \\
&  =\prod_{t_{i}^{(k)}\leq t}\left \{  1-\frac{\alpha(\mathbf{x},\mathbf{x}%
_{i})}{1-\sum_{j=1}^{i-1}\alpha(\mathbf{x},\mathbf{x}_{j})}\right \}
^{\delta_{i}}.
\end{align}

By having $M$ random subsets $\mathcal{A}_{k}$, $k=1,...,M$, we can adapt the
bagging method to the Beran estimators as base models and construct an
aggregated SF $S(t\mid \mathbf{x})$ as the mean values of SFs predicted by the
Beran estimators:
\begin{equation}
S(t\mid \mathbf{x})=\frac{1}{M}\sum_{k=1}^{M}S^{(k)}(t\mid \mathbf{x}).
\end{equation}

This is the simplest way to implement the ensemble of Beran estimators.
However, a more flexible aggregation of the Beran estimators as base models
can be proposed.

\section{SurvBETA: Attention-based ensemble}

One of the ways to improve the aggregation scheme in ensemble-based models is
to introduce some weights $\gamma_{i}$, $i=1,...,M$, for the weak model
predictions. The weights should take into account how an analyzed point
$\mathbf{x}$ is far from other points in the training set. Suppose that the
training set $\mathcal{A}$ consists of two clusters such that two bootstrap
samples with indices $i$ and $j$ mainly coincide with these clusters. Let
$\mathbf{x}$ belong to the $i$-th cluster. The $i$-th Beran estimator trained
on $\mathcal{A}_{i}$ recognizes $\mathbf{x}$ as an instance from the first
cluster because it does not \textquotedblleft know\textquotedblright \ the
second cluster. It provides the prediction $S^{(i)}(t\mid \mathbf{x})$. The
$j$-th Beran estimator trained on $\mathcal{A}_{j}$ may provide a similar
estimation of the SF $S^{(j)}(t\mid \mathbf{x})$ because the corresponding
weights are normalized and are not sensitive to the considered case.
Therefore, new weights assigned to predictions and taking into account the
distance between the analyzed feature vector $\mathbf{x}$ and all subsets
$\mathcal{A}_{k}$, $k=1,...,M$, could overcome the above difficulty. Similar
approach has been used \cite{Utkin-Konstantinov-22,Utkin-Konstantinov-22d}
where the random forests and random survival forest were considered as
ensemble-based models.

\subsection{Prototypes of subsets}

Another problem is how to define the distance between the analyzed vector
$\mathbf{x}$ and the subset $\mathcal{A}_{k}$. In
\cite{Utkin-Konstantinov-22,Utkin-Konstantinov-22d}, the mean value
$\mathbf{e}(\mathcal{A}_{k})$ of all feature vectors from $\mathcal{A}_{k}$
was used to measure the distance, i.e. the distance between $\mathbf{x}$ and
$\mathbf{e}(\mathcal{A}_{k})$. At that, in contrast to SurvBETA the set
$\mathcal{A}_{k}$ in \cite{Utkin-Konstantinov-22,Utkin-Konstantinov-22d} was
taken from leaves of trees. The mean value can be regarded as a prototype of
instances from $\mathcal{A}_{k}$. However, the mean value can also lead to
some difficulties when the bootstrap sample consists from instances belonging
to different clusters. In this case, the corresponding prototype may be
incorrect. Therefore, we propose to use the attention mechanism to find the
prototype relative to the vector $\mathbf{x}$ such that the attention weights
depends on distances between $\mathbf{x}$ and feature vectors from
$\mathcal{A}_{k}$.

Formally, the mean value $\mathbf{e}(\mathcal{A}_{k})$ of the feature vectors
from the subset $\mathcal{A}_{k}$ is defined as:
\begin{equation}
\mathbf{e}(\mathcal{A}_{k})=\frac{1}{\# \mathcal{A}_{k}}\sum_{\mathbf{x}%
_{j}\in \mathcal{A}_{k}}\mathbf{x}_{j},
\end{equation}
where $\# \mathcal{A}_{k}$ denotes the number of instances in $\mathcal{A}%
_{k}$.

The above mean value can be regarded as a prototype of the subset
$\mathcal{A}_{k}$, but this prototype does not depend on the analyzed vector
$\mathbf{x}$. Therefore, it is proposed another prototype which relates the
vector $\mathbf{x}$. The prototype vector of $\mathcal{A}_{k}$ relative to
$\mathbf{x}$ can be written by using the Nadaraya-Watson regression as
follows:
\begin{equation}
\mathbf{e}(\mathcal{A}_{k},\mathbf{x})=\sum_{\mathbf{x}_{j}\in \mathcal{A}_{k}%
}\mu_{k}(\mathbf{x},\mathbf{x}_{j})\cdot \mathbf{x}_{j},
\end{equation}
where $\mu_{k}(\mathbf{x},\mathbf{x}_{j})$ is the attention weight
establishing the relationship between $\mathbf{x}$ and each element of
$\mathcal{A}_{k}$ such that
\begin{equation}
\mu_{k}(\mathbf{x},\mathbf{x}_{j})=\text{\textrm{softmax}}\left(
-\frac{\left \Vert \mathbf{x}-\mathbf{x}_{j}\right \Vert ^{2}}{\eta_{k}}\right)
. \label{atten_prot1}%
\end{equation}

Here $\eta_{k}$ is a hyperparameter or the training parameter depending on the
implementation strategy. To simplify calculations, we can learn only one
parameter $\eta$ for all base learners.

The first definition of the prototype $\mathbf{e}(\mathcal{A}_{k})$ does not
depend on $\mathbf{x}$ and can be calculated in advance for all models in the
ensemble and then used for any $\mathbf{x}$. This reduces calculations.
However, vectors in $\mathcal{A}_{k}$ may be too different, for example, they
may be from different clusters. This fact leads to a significant bias of the
prototype. Therefore, the second definition is preferable when the data
structure is known to be complex.

\subsection{A specific scheme of generating subsets}

Let us analyze how a prototype $\mathbf{e}(\mathcal{A}_{k},\mathbf{x})$ is
located in $\mathcal{A}$. Since $\mathbf{e}(\mathcal{A}_{k},\mathbf{x})$ is a
convex combination of all $\mathbf{x}_{j}$ from $\mathcal{A}_{k}$, then the
prototype is located inside the subset $\mathcal{A}_{k}$. Moreover,
contribution of points $\mathbf{x}_{j}$ which are close to $\mathbf{x}$ is
large due to the kernel definition of weights $\mu_{k}(\mathbf{x}%
,\mathbf{x}_{j})$. Hence, we can conclude that the prototype $\mathbf{e}%
(\mathcal{A}_{k},\mathbf{x})$ is close to $\mathbf{x}$ if $\mathbf{x}$ is
inside the subset $\mathcal{A}_{k}$. On the other hand, the prototype
$\mathbf{e}(\mathcal{A}_{k},\mathbf{x})$ may be far from $\mathbf{x}$ if
$\mathbf{x}$ is outside the subset $\mathcal{A}_{k}$. As a result, if
bootstrap samples $\mathcal{A}_{k}$ and $\mathcal{A}_{l}$ are far from each
other or just different and $\mathbf{x}$ is outside the subset $\mathcal{A}%
_{k}$, then the weight $\gamma \left(  \mathbf{x},\mathbf{e}(\mathcal{A}%
_{k},\mathbf{x})\right)  $ of the $k$-th Beran estimator will be very large
whereas the weight $\gamma \left(  \mathbf{x},\mathbf{e}(\mathcal{A}%
_{l},\mathbf{x})\right)  $ will be very small. This implies that the ensemble
does not improve the prediction accuracy in comparison with a single Beran
estimator. In contrast to many ensemble models, for example, random forests,
the subsets $\mathcal{A}_{k}$, $k=1,...,M$, should be intersected. In this
case, if $\mathbf{x}$ is located in the intersection area of two samples, then
the corresponding weights, say $\gamma \left(  \mathbf{x},\mathbf{e}%
(\mathcal{A}_{k},\mathbf{x})\right)  $ and $\gamma \left(  \mathbf{x}%
,\mathbf{e}(\mathcal{A}_{l},\mathbf{x})\right)  $, will have comparable values
and the corresponding Beran estimators significantly contribute into the final
estimate. This is important property of the proposed ensemble. On the one
hand, we aim to achieve independence of subsamples. On the other hand, this
independence may lead to the ensemble degradation. Therefore, we propose the
following procedure for generating subsets $\mathcal{A}_{k}$.

\begin{enumerate}
\item $M$ points $\mathbf{x}$ from the training set $\mathcal{A}$ are randomly selected.

\item $K$ nearest neighbors for the $k$-th randomly selected point are taken
from $\mathcal{A}$ to construct the subset $\mathcal{A}_{k}$, $k=1,...,M$.

\item The obtained subsets $\mathcal{A}_{k}$, $k=1,...,M$, are specific
bootstrap samples.
\end{enumerate}

Here $K$ can be regarded as a hyperparameter. It is determined based on the
fact that subsets should have certain areas of intersection.

\subsection{Global attention}

Let us apply the attention mechanism to computing the ensemble-based
prediction of the Beran estimators. By using notation introduced in the
previous sections, we can write
\begin{equation}
S(t\mid \mathbf{x})=\sum_{k=1}^{M}\gamma \left(  \mathbf{x},\mathbf{e}%
(\mathcal{A}_{k},\mathbf{x})\right)  \cdot S^{(k)}(t\mid \mathbf{x}),
\label{Ens_Ber_36}%
\end{equation}
where the attention weight $\gamma$ is determined as
\begin{equation}
\gamma \left(  \mathbf{x},\mathbf{e}(\mathcal{A}_{k},\mathbf{x})\right)
=\text{\textrm{softmax}}\left(  -\frac{\left \Vert \mathbf{x}-\mathbf{e}%
(\mathcal{A}_{k},\mathbf{x})\right \Vert ^{2}}{w_{k}}\right)  ,
\end{equation}
where $w_{k}$ is the training parameter.

\begin{figure}
[ptb]
\begin{center}
\includegraphics[
height=5.0in,
width=5.0in
]%
{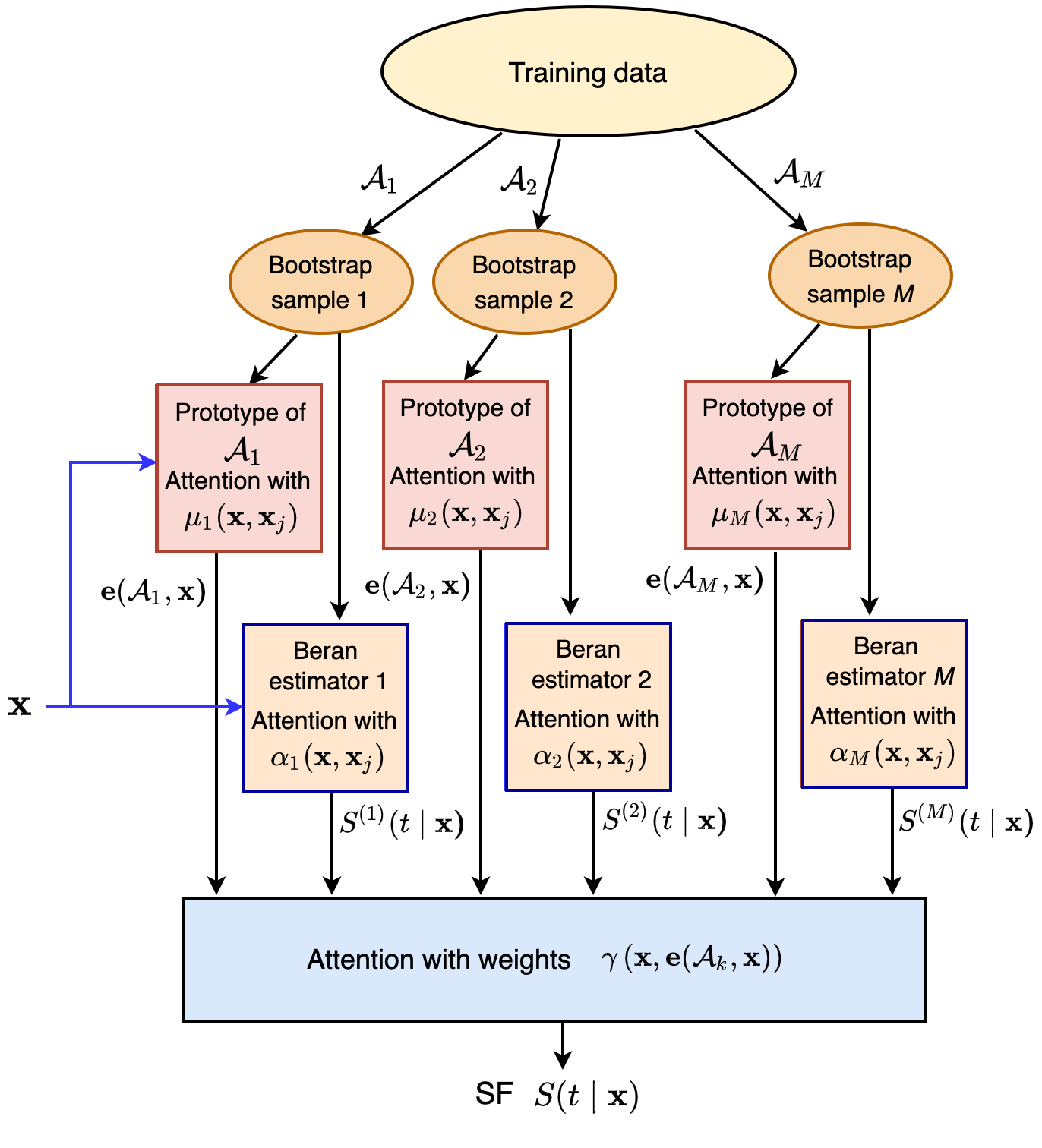}%
\caption{A structure of attention mechanisms with the corresponding attention
weights}%
\label{f:ens_ber_1}%
\end{center}
\end{figure}

The considered structure of the above attention mechanisms with the
corresponding attention weights, including the Beran estimator weights, is
depicted in Fig. \ref{f:ens_ber_1}.%

Let us write for short:
\begin{equation}
\gamma_{i}^{(k)}=\gamma \left(  \mathbf{x}_{i},\mathbf{e}(\mathcal{A}%
_{k},\mathbf{x}_{i})\right)  .
\end{equation}

In sum, we have three type of attention weights: first, weights in the Beran
estimator denoted as $\alpha(\mathbf{x},\mathbf{x}_{j})$; second, weights for
computing the prototype $\mathbf{e}(\mathcal{A}_{k},\mathbf{x})$ of
$\mathcal{A}_{k}$ for $\mathbf{x}$ denoted as $\mu_{k}(\mathbf{x}%
,\mathbf{x}_{j})$; third, weights for aggregating the Beran estimator
predictions $S^{(k)}(t\mid \mathbf{x})$ denoted as $\gamma \left(
\mathbf{x},\mathbf{e}(\mathcal{A}_{k},\mathbf{x})\right)  $. The above weights
can be training as well as tuning.

\subsection{Loss functions}

By supposing that SFs are aggregated in accordance with (\ref{Ens_Ber_36}), it
is obvious that the expected event time $\widehat{T}_{i}$ of $\mathbf{x}_{i}$
for the ensemble is computed as weighted sum of expected event times
$\widehat{T}_{i}^{(k)}$ of $\mathbf{x}_{i}$ predicted by weak models:%
\begin{equation}
\widehat{T}_{i}=\sum_{k=1}^{M}\gamma_{i}^{(k)}\cdot \widehat{T}_{i}^{(k)}.
\end{equation}

Finally, parameters $w_{k}$ and $\eta_{k}$, $k=1,...,M$, can be found by
minimizing the C-index loss $L^{C}$ written in terms of expected times:%
\begin{align}
\min_{\substack{\eta_{k},w_{k},\\k=1,...,M}}L^{C}  &  =\min_{\substack{\eta
_{k},w_{k},\\k=1,...,M}}\frac{1}{N}\sum_{(i,j)\in \mathcal{J}}\mathbf{1}\left[
\widehat{T}_{i}-\widehat{T}_{j}>0\right] \nonumber \\
&  =\min_{\substack{\eta_{k},w_{k},\\k=1,...,M}}\sum_{(i,j)\in \mathcal{J}%
}\mathbf{1}\left[  \sum_{k=1}^{M}\left(  \gamma_{i}^{(k)}\widehat{T}_{i}%
^{(k)}-\gamma_{j}^{(k)}\widehat{T}_{j}^{(k)}\right)  >0\right]  .
\label{object_f_1}%
\end{align}

Since we have a finite number of events, then the SF is step-wised, then we
can write for the $k$-th Beran estimator%
\begin{equation}
S^{(k)}(t\mid \mathbf{x}_{i})=\sum \limits_{l=0}^{N_{k}-1}S_{l}^{(k)}%
(\mathbf{x}_{i})\cdot \mathbf{1}\{t\in \lbrack t_{l}^{(k)},t_{l+1}^{(k)})\},
\end{equation}
where $S_{l}^{(k)}(\mathbf{x}_{i})=S^{(k)}(t_{l}\mid \mathbf{x}_{i})$ is the SF
in the time interval $[t_{l}^{(k)},t_{l+1}^{(k)})$; $S_{0}^{(k)}=1$
by\ $t_{0}=0$; $N_{k}$ is the number of elements in $\mathcal{A}_{k}$;
$t_{1}^{(k)}<t_{2}^{(k)}<...t_{N_{k}}^{(k)}$ are ordered times to events from
$\mathcal{A}_{k}$.

Hence, the expected time $\widehat{T}_{i}^{(k)}$ is determined as
\begin{equation}
\widehat{T}_{i}^{(k)}=\sum \limits_{l=0}^{N_{k}-1}S_{l}^{(k)}(\mathbf{x}%
_{i})(t_{l+1}^{(k)}-t_{l}^{(k)}).
\end{equation}

Then the optimization problem (\ref{object_f_1}) is rewritten as
\begin{equation}
\min_{\substack{\eta_{k},w_{k},\\k=1,...,M}}L^{C}=\min_{\substack{\eta
_{k},w_{k},\\k=1,...,M}}\sum_{(i,j)\in \mathcal{J}}\mathbf{1}\left[
G_{ij}>0\right]  , \label{Ens_Ber_40}%
\end{equation}
where
\begin{align}
G_{ij}  &  =\sum_{k=1}^{M}\gamma_{i}^{(k)}\sum \limits_{l=0}^{N_{k}-1}%
S_{l}^{(k)}(\mathbf{x}_{i})(t_{l+1}^{(k)}-t_{l}^{(k)})\nonumber \\
&  -\sum_{k=1}^{M}\gamma_{j}^{(k)}\sum \limits_{l=0}^{N_{k}-1}S_{l}%
^{(k)}(\mathbf{x}_{j})(t_{l+1}^{(k)}-t_{l}^{(k)}).
\end{align}

It should be noted that the indicator function in the C-index can be replaced
with the sigmoid function $\sigma$ to simplify the optimization problem.

The C-index has been used above as a loss function. A more complex loss
function was proposed in \cite{Hu-etal-21}. Besides the C-index, it the loss
function additionally contains several terms. It additionally contains several
terms: $L^{obs}$, $L^{cen}$, $L^{MAE}$. The first term is determined only for
uncensored observations as follows:
\begin{equation}
L^{obs}=-\sum_{i\in \mathcal{C}}\sum_{t_{j}\leq T_{i}}\ln S(t_{j}\mid
\mathbf{x}_{i})-\sum_{i\in \mathcal{C}}\sum_{t_{j}>T_{i}}\ln \left(
1-S(t_{j}\mid \mathbf{x}_{i})\right)  .
\end{equation}

Here $\mathcal{C}\subseteq \mathcal{A}$ is the set of all uncensored
observation indices. The loss $L^{obs}$ maximizes the SF $S(t_{j}%
\mid \mathbf{x}_{i})$ for $t_{j}\leq T_{i}$ and minimizes it for $t_{j}>T_{i}$.

The second term is determined for censored observations whose index set
denoted as $\mathcal{U}=\mathcal{A}\backslash \mathcal{C}$. It is of the form:
\begin{equation}
L^{cen}=-\sum_{i\in \mathcal{U}}\sum_{t_{j}\leq T_{i}}\ln S(t_{j}\mid
\mathbf{x}_{i}).
\end{equation}

The third term is determined as
\begin{equation}
L^{MAE}=\sum_{i\in \mathcal{C}}\left \vert T_{i}-\widehat{T}_{i}\right \vert
=\sum_{i\in \mathcal{C}}\left \vert T_{i}-\sum \limits_{l=1}^{m}S_{l}%
(\mathbf{x}_{i})(t_{l+1}-t_{l})\right \vert . \label{L_MAE}%
\end{equation}

\subsection{General case}

Generally, the above attention weights can be represented through the scaled
dot product as follows:
\begin{equation}
\alpha(\mathbf{x},\mathbf{x}_{i})=\mathrm{softmax}\left(  \frac{\left(
\mathbf{q}_{\alpha}\otimes \mathbf{x}\right)  \mathbf{\cdot}\left(
\mathbf{k}_{\alpha}\otimes \mathbf{x}_{i}\right)  }{\sqrt{d_{\mathbf{x}}}%
}\right)  , \label{alfa_gen}%
\end{equation}%
\begin{equation}
\mu_{k}(\mathbf{x},\mathbf{x}_{j})=\text{\textrm{softmax}}\frac{\left(
\left(  \mathbf{q}_{\mu}\otimes \mathbf{x}\right)  \mathbf{\cdot}\left(
\mathbf{k}_{\mu}\otimes \mathbf{x}_{j}\right)  \right)  }{\sqrt{d_{\mathbf{x}}%
}}, \label{mu_gen}%
\end{equation}%
\begin{equation}
\gamma \left(  \mathbf{x},\mathbf{e}(\mathcal{A}_{k},\mathbf{x})\right)
=\text{\textrm{softmax}}\left(  \frac{\left(  \mathbf{q}_{\gamma}%
\otimes \mathbf{x}\right)  \mathbf{\cdot}\left(  \mathbf{k}_{\gamma}%
\otimes \mathbf{x}_{j}\right)  }{\sqrt{d_{\mathbf{x}}}}\right)  .
\label{gamma_gen}%
\end{equation}

Here $\mathbf{q}_{\alpha}$, $\mathbf{q}_{\mu}$, $\mathbf{q}_{\gamma}%
\in \mathbb{R}^{d}$ are trainable vectors of query parameters; $\mathbf{k}%
_{\alpha}$, $\mathbf{k}_{\mu}$, $\mathbf{k}_{\gamma}\in \mathbb{R}^{d}$ are
trainable vectors of key parameters; $\otimes$ is the symbol of the Hadamard
product of vectors; $d_{\mathbf{x}}$ is the dimensionality of the vector
$\mathbf{x}$.

Replacing the indicator function with the sigmoid function $\sigma
(z,\tau)=1/(1+\exp(-\tau z))$, the optimization problem (\ref{Ens_Ber_40}) is
represented as
\begin{equation}
\min_{\substack{\mathbf{q}_{\alpha},\mathbf{q}_{\mu},\mathbf{q}_{\gamma
},\\ \mathbf{k}_{\alpha},\mathbf{k}_{\mu},\mathbf{k}_{\gamma}}}\sum
_{(i,j)\in \mathcal{J}}\frac{1}{1+\exp(\tau G_{ij})}. \label{general_problem}%
\end{equation}

The parameter $\tau$ should be larger $1$ to ensure proximity of the sigmoid
function to the indicator function.

A main difficulty for implementing the general case is the large number of
training parameters which lead to overfitting especially when the model is
trained on small datasets. Therefore, we propose an important special case
which significantly simplifies the optimization problem and reduces the number
of training parameters.

\subsection{Important special case\label{subsec:import_case}}

If we suppose that $\eta_{k}$, $k=1,...,M$, in (\ref{atten_prot1}) are
hyperparameters which can be identical for all base learners, but are
generally different, then the optimization problem for computing parameters
$v_{k}$, $k=1,...,M,$ in (\ref{Ens_Ber_40}) can significantly be simplified.
Let us use the definition of the attention weight $\gamma^{(k)}=\gamma \left(
\mathbf{x},\mathbf{e}(\mathcal{A}_{k},\mathbf{x})\right)  $ in the following
form proposed in \cite{Utkin-Konstantinov-22}:%
\begin{align}
\gamma_{i}^{(k)}  &  =(1-\epsilon)\cdot \text{\textrm{softmax}}\left(
-\frac{\left \Vert \mathbf{x}_{i}-\mathbf{e}(\mathcal{A}_{k},\mathbf{x}%
_{i})\right \Vert ^{2}}{w}\right)  +\epsilon \cdot v_{k},\nonumber \\
k  &  =1,...,M. \label{Gamma}%
\end{align}

This attention weight is derived from the imprecise Huber's $\epsilon
$-contamination model \cite{Huber81} which is represented as
\begin{equation}
F=(1-\epsilon)\cdot P+\epsilon \cdot V,
\end{equation}
where $P$ is a discrete probability distribution contaminated by another
probability distribution $V$ that is arbitrary in the unit simplex $\Delta
^{M}$ such that conditions $v_{1}+...+v_{M}=1$ and $v_{k}\geq0$ for all
$k=1,...,M$ are valid for components of $V$; the contamination parameter
$\epsilon \in \lbrack0,1]$ controls the size of the small simplex produced by
the $\epsilon$-contamination model. In particular, if $\epsilon=0$, then the
model is reduced to the softmax operation with hyperparameter $w$, and the
attention weight does not depend on $v_{k}$ in this case. The distribution $P$
has elements \textrm{softmax}$\left(  -\left \Vert \mathbf{x}-\mathbf{e}%
(\mathcal{A}_{k},\mathbf{x})\right \Vert ^{2}/w\right)  $, $k=1,...,M$. The
distribution $V$ defines the vector $\mathbf{v}=(v_{1},...,v_{M})$ of training
parameter of the attention mechanism.

The next step to simplify the optimization problem for computing parameters
$v_{k}$ is to replace the indicator function in (\ref{Ens_Ber_40}) with the
hinge loss function $l(x)=\max \left(  0,x\right)  $ similarly to the
replacement proposed by Van Belle et al. \cite{Van_Belle-etal-2007}.

By adding the regularization term, the optimization problem can be written as%
\begin{equation}
\min_{\mathbf{v}\in \Delta^{M}}\left \{  \sum_{(i,j)\in \mathcal{J}}\max \left(
0,G_{ij}\right)  \right \}  . \label{Survival_DF_46}%
\end{equation}

Let us introduce the variables%
\begin{equation}
\xi_{ij}=\max \left(  0,G_{ij}\right)  . \label{Survival_DF_48}%
\end{equation}

Then the optimization problem can be written as follows:
\begin{equation}
\min_{\mathbf{v,}\xi_{ij}}\sum_{(i,j)\in \mathcal{J}}\xi_{ij},
\label{Survival_DF_50}%
\end{equation}
subject to $\mathbf{v}\in \Delta^{M}$ and $\xi_{ij}\geq G_{ij}$, $\xi_{ij}%
\geq0$, $\{i,j\} \in \mathcal{J}$.

Denote%
\begin{equation}
P_{i}^{(k)}=\mathrm{softmax}\left(  -\frac{\left \Vert \mathbf{x}%
_{i}-\mathbf{e}(\mathcal{A}_{k},\mathbf{x}_{i})\right \Vert ^{2}}{w}\right)  ,
\end{equation}%
\begin{align}
Q_{ij}  &  =(1-\epsilon)\sum_{k=1}^{M}P_{i}^{(k)}\sum \limits_{l=0}^{N_{k}%
-1}S_{l}^{(k)}(\mathbf{x}_{i})(t_{l+1}^{(k)}-t_{l}^{(k)})\nonumber \\
&  -(1-\epsilon)\sum_{k=1}^{M}P_{j}^{(k)}\sum \limits_{l=0}^{N_{k}-1}%
S_{l}^{(k)}(\mathbf{x}_{j})(t_{l+1}^{(k)}-t_{l}^{(k)}),
\end{align}%
\begin{equation}
R_{ij}^{(k)}=\left(  \sum \limits_{l=0}^{N_{k}-1}S_{l}^{(k)}(\mathbf{x}%
_{i})(t_{l+1}^{(k)}-t_{l}^{(k)})-\sum \limits_{l=0}^{N_{k}-1}S_{l}%
^{(k)}(\mathbf{x}_{j})(t_{l+1}^{(k)}-t_{l}^{(k)})\right)  .
\end{equation}

Then we obtain the linear optimization problem:
\begin{equation}
\min_{\mathbf{v,}\xi_{ij}}\sum_{(i,j)\in \mathcal{J}}\xi_{ij},
\label{spec_case_1}%
\end{equation}
subject to $\mathbf{v}\in \Delta^{M}$ and
\begin{equation}
\xi_{ij}-\epsilon \sum_{k=1}^{M}R_{ij}^{(k)}v_{k}\geq Q_{ij},\  \  \xi_{ij}%
\geq0,\  \  \{i,j\} \in \mathcal{J}. \label{spec_case_2}%
\end{equation}

The above problem has $M$ variables $v_{1},...,v_{M}$ and $\# \mathcal{J}$
variables $\xi_{ij}$.

Let us extend the loss function by adding the term $L^{MAE}$ defined in
(\ref{L_MAE}). The predicted mean value $\widehat{T}_{i}$ is determined
through the predicted mean times $\widehat{T}_{i}^{(k)}$ of each Beran
estimator as
\begin{equation}
\widehat{T}_{i}=\sum_{k=1}^{M}\gamma_{i}^{(k)}\cdot \widehat{T}_{i}^{(k)}%
=\sum_{k=1}^{M}\left(  (1-\epsilon)P_{i}^{(k)}\widehat{T}_{i}^{(k)}%
+\epsilon \cdot v_{k}\widehat{T}_{i}^{(k)}\right)  .
\end{equation}

Denote
\begin{equation}
U_{i}^{(k)}=(1-\epsilon)P_{i}^{(k)}\widehat{T}_{i}^{(k)}.
\end{equation}

Then the extended objective function (\ref{spec_case_1}) becomes
\begin{equation}
\sum_{(i,j)\in \mathcal{J}}\xi_{ij}+\sum_{i\in \mathcal{C}}\left \vert
T_{i}-\widehat{T}_{i}\right \vert =\sum_{(i,j)\in \mathcal{J}}\xi_{ij}%
+\sum_{i\in \mathcal{C}}\left \vert \sum_{k=1}^{M}\left(  U_{i}^{(k)}%
+\epsilon \widehat{T}_{i}^{(k)}\cdot v_{k}\right)  -T_{i}\right \vert .
\end{equation}

Let us introduce new optimization variables $\psi_{i}$, $i\in \mathcal{C}$,
such that
\begin{equation}
\psi_{j}=\left \vert \sum_{k=1}^{M}\left(  U_{i}^{(k)}+\epsilon \widehat{T}%
_{i}^{(k)}\cdot v_{k}\right)  -T_{i}\right \vert .
\end{equation}

Then we obtain the following optimization problem:
\begin{equation}
\min_{\mathbf{v,}\xi_{ij},\psi_{i}}\left(  \sum_{(i,j)\in \mathcal{J}}\xi
_{ij}+\sum_{i\in \mathcal{C}}\psi_{i}\right)  , \label{C-ind_MAE}%
\end{equation}
subject to $\mathbf{v}\in \Delta^{M}$ and
\begin{equation}
\xi_{ij}-\epsilon \sum_{k=1}^{M}R_{ij}^{(k)}v_{k}\geq Q_{ij},\  \  \xi_{ij}%
\geq0,\  \  \{i,j\} \in \mathcal{J},
\end{equation}%
\begin{align}
\psi_{i}-\epsilon \sum_{k=1}^{M}\widehat{T}_{i}^{(k)}v_{k}  &  \geq \sum
_{k=1}^{M}U_{i}^{(k)}-T_{i},\ i\in \mathcal{C},\nonumber \\
\psi_{i}+\epsilon \sum_{k=1}^{M}\widehat{T}_{i}^{(k)}v_{k}  &  \geq T_{i}%
-\sum_{k=1}^{M}U_{i}^{(k)},\ i\in \mathcal{C}.
\end{align}

We again have the linear optimization problem with additional variables
$\psi_{i}$, $i\in \mathcal{C}$.

The optimization problem (\ref{spec_case_1})-(\ref{spec_case_2}) can be also
represented as a quadratic programming problem if to add the regularization
term $\left \Vert \mathbf{v}\right \Vert ^{2}$ with the hyperparameter $\lambda$
controlling the strength of the regularization. In this case, we obtain the
objective function
\begin{equation}
\min_{\mathbf{v,}\xi_{ij},\psi_{i},\epsilon}\left(  \sum_{(i,j)\in \mathcal{J}%
}\xi_{ij}+\sum_{i\in \mathcal{C}}\psi_{i}+\lambda \left \Vert \mathbf{v}%
\right \Vert ^{2}\right)  . \label{special_problem}%
\end{equation}

The parameter $\epsilon$ can be considered as the trainable parameter if to
introduce new variables $\beta_{k}=\epsilon v_{k}$. This replacement leads to
new constraints
\begin{equation}
\sum_{k=1}^{M}\beta_{k}=\epsilon,\  \  \beta_{k}\geq0,\ k=1,...,M.
\label{new_constr}%
\end{equation}

Then the above optimization problem is rewritten as
\begin{equation}
\min_{\mathbf{v,}\xi_{ij},\psi_{i},\beta_{k}}\left(  \sum_{(i,j)\in
\mathcal{J}}\xi_{ij}+\sum_{i\in \mathcal{C}}\psi_{i}\right)  ,
\end{equation}
subject to (\ref{new_constr}) and
\begin{equation}
\xi_{ij}-\sum_{k=1}^{M}R_{ij}^{(k)}\beta_{k}\geq Q_{ij},\  \  \xi_{ij}%
\geq0,\  \  \{i,j\} \in \mathcal{J},\  \epsilon \geq0,\  \  \epsilon \leq1,
\end{equation}%
\begin{align}
\psi_{i}-\sum_{k=1}^{M}\widehat{T}_{i}^{(k)}\beta_{k}  &  \geq \sum_{k=1}%
^{M}U_{i}^{(k)}-T_{i},\ i\in \mathcal{C},\nonumber \\
\psi_{i}+\sum_{k=1}^{M}\widehat{T}_{i}^{(k)}\beta_{k}  &  \geq T_{i}%
-\sum_{k=1}^{M}U_{i}^{(k)},\ i\in \mathcal{C}.
\end{align}

\subsection{Computational problems}

The above model has an important obstacle. When a training set is rather
large, numbers of variables in (\ref{general_problem}) and
(\ref{special_problem}) as well as constraints in (\ref{special_problem}) are
extremely large. This is a common computational problem of survival machine
learning problems where the loss function uses the C-index. To simplify the
problem, Hu et al. \cite{Hu-etal-21} proposed to apply a simplified randomized
algorithm. According to the algorithm, for each observed object $\mathbf{x}%
_{i}$ in the training set, we randomly sample another object $\mathbf{x}_{j}$
is randomly sampled with replacement, where $T_{i}<T_{j}$.

A nearest neighbor algorithm to reduce the computational complexity was
proposed in \cite{Belle-etal-2008}. According to this algorithm, the number of
constraints can be reduced by selecting a subset $\mathcal{J}_{i}$ of $k$
objects with a survival time nearest to the survival time of the object
$\mathbf{x}_{i}$. Another approach presented in \cite{Utkin-etal-2019c} is to
reduce the number of pairs in $\mathcal{J}$ by random selection of $K$ pairs.

\section{Numerical experiments}

In numerical experiments, we study properties the proposed models SurvBETA
trained on synthetic and real data and compare them with other available
survival models. A performance measure for studying and comparing models is
the C-index computed on the testing set. To evaluate the C-index for each
dataset, we perform a cross-validation with $100$ repetitions by taking 20\% of instances from each dataset for testing, 60\%
of instances for training and 20\% of instances for validation.
Instances for training, validation and testing are randomly selected in each run.
Hyperparameters are optimized by using the Optuna library
\cite{akiba2019optuna}. For SurvBETA without optimization and the Beran
estimator, the complete enumeration of all hyperparameter values in predefined
intervals is performed.

In numerical experiments, three models are considered which can be regarded as
special cases of SurvBETA. The first model is just a single Beran estimator.
It is a special case of SurvBETA when the number of estimators equal to $1$.
This case is mainly used for comparison purposes. Another special case is when
all parameters of the SurvBETA are tuning, i.e., they are determined by means
of testing. Actually, the second special case corresponds to computing the
attention weights $\gamma_{i}^{(k)}$ in (\ref{Gamma}) under condition that
$\epsilon=0$. This case will be called \textquotedblleft SurvBETA without
optimization\textquotedblright \ because the corresponding modification of
SurvBETA does not use any optimization problem for computing parameters. The
third special case of SurvBETA is based on applying the optimization problem
(\ref{C-ind_MAE}) for computing optimal values of training variables
$\mathbf{v}=(v_{1},...,v_{M})$ in (\ref{Gamma}). This case will be called
\textquotedblleft SurvBETA with optimization\textquotedblright.

In all experiments, four standard kernels are used for computing the Beran
estimator and for the prototype computation: the Gaussian kernel, the
Epanechnikov kernel $K(\mathbf{x},\mathbf{x}_{i})=\frac{3}{4}(1-u^{2})$, the
triangular kernel $K(\mathbf{x},\mathbf{x}_{i})=(1-\left \vert u\right \vert )$,
and the quartic kernel $K(\mathbf{x},\mathbf{x}_{i})=\frac{15}{16}%
(1-u^{2})^{2}$, where $u=-\left \Vert \mathbf{x}-\mathbf{x}_{i}\right \Vert
^{2}/\tau;$ $\tau$ is the hyperparameter which is selected from the set ${10^{-3},10^{-2},10^{-1},10^{0},10^{1},10^{2},10^{3}}$ 
for each weak learner. For every subsample, one kernel is
randomly selected from the above four kernels, and it is used in the
corresponding Beran estimator and in the attention mechanism for determining
the prototype $\mathbf{e}(\mathcal{A}_{k},\mathbf{x})$. This approach allows
us to make the weak models in the ensemble more diverse. Weights
$\gamma \left(  \mathbf{x},\mathbf{e}(\mathcal{A}_{k},\mathbf{x})\right)  $ use
only the Gaussian kernels.

It is important to note that we do not consider the problem
(\ref{general_problem}) with attention weights in general forms
(\ref{alfa_gen})-(\ref{gamma_gen}) because the large number of training
parameters (vectors $\mathbf{q}_{\alpha}$, $\mathbf{q}_{\mu}$, $\mathbf{q}%
_{\gamma}$, $\mathbf{k}_{\alpha}$, $\mathbf{k}_{\mu}$, $\mathbf{k}_{\gamma}$)
leads to overfitting when the number of instance is restricted. Therefore, we
consider only the important special case which is provided in Subsection
\ref{subsec:import_case}.

\subsection{Synthetic data}

Two clusters of covariates $\mathbf{x}\in \mathbb{R}^{d}$ ($d=5$) are randomly
generated such that points of every cluster are generated from the uniform
distribution in a hyper-rectangle $\prod_{j=1}^{d}[l_{i}^{(j)},r_{i}^{(j)}]$,
$i=1,2$. The number of points denoted as $N$ in the clusters varies to study
its impact on the model performance. The censored data are generated randomly
according to the Bernoulli distribution with probabilities $\Pr \{
\delta=0\}=0.2$ and $\Pr \{ \delta_{i}=1\}=0.8$. Event times are generated in
accordance with the Weibull distribution with the shape parameter $k$, which
depends on one of the dimensions $x_{1}$ and has the following form:%

\begin{equation}
T=\frac{\sin(c\cdot x_{1})+c}{\Gamma \left(  1+\frac{1}{k}\right)  }%
\cdot \left(  -\log(u)\right)  ^{\frac{1}{k}},
\end{equation}
where $c$ is a parameter; $u$ is a random variable uniformly distributed in
the interval $[0,1]$.

In order to complicate synthetic data, we take the expectation of $T$ in
accordance with the Weibull distribution as $\sin(c\cdot
x_{1})+c$. In other words, the
mean value of the event time is changed as the sinusoid function.

Intervals of hyperparameter values for different models are the following:

\begin{enumerate}
\item SurvBETA with optimization: $\epsilon \in \lbrack0,1]$; $w\in
\{10^{-3},10^{-2},10^{-1},10^{0},10^{1},10^{2},10^{3}\}$.

\item SurvBETA without optimization of weights: $w\in \{10^{-3},10^{-2}%
,10^{-1},10^{0},10^{1},10^{2},10^{3}\}$.

\item The single Beran estimator: $\tau \in \{10^{-3},10^{-2},10^{-1}%
,10^{0},10^{1},10^{2},10^{3}\}$.
\end{enumerate}

These hyperparameters are determined by using the Optuna library
\cite{akiba2019optuna}.  Other hyperparameters are manually tested, choosing those leading to the best results.

In order to investigate how the performance results (the C-index) depend on
the different parameters, the corresponding experiments are performed. In all
experiments, training parameters are optimized by using Adam with the initial
learning rate $0.1$ and $60$ epochs. Initial parameters of synthetic data are
$c=3$, $k=6$, $l_{1}^{(j)}=-2.0$, $r_{1}^{(j)}=l_{1}^{(j)}+4$, $l_{2}%
^{(j)}=20.0$, $r_{2}^{(j)}=l_{2}^{(j)}+10$, $j=1,...,5$. The initial number of
points in every cluster is $500$. The initial size of each bootstrap subsample
for constructing the ensemble is $40\%$. The number of weak models is $20$. Some of the above parameters are
changed in experiments to study their impact on the resulting C-index.

\begin{figure}
[ptb]
\begin{center}
\includegraphics[
height=2.8778in,
width=3.8291in
]%
{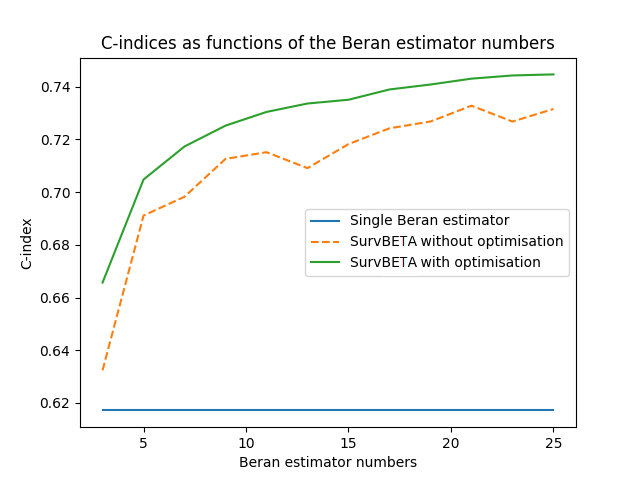}%
\caption{Dependence of the C-index on the number of the Beran estimators}%
\label{fig:estimator_dependency}%
\end{center}
\end{figure}

\emph{Dependence of the C-index on the number of the Beran estimators in the
ensemble:} the number of estimators varies in the range $[3,26]$ with step
$2$. The corresponding dependence is shown in Fig.
\ref{fig:estimator_dependency}. It can be seen from Fig.
\ref{fig:estimator_dependency} that the C-index of SurvBETA without
optimization as well as SurvBETA with optimization increases as the number of
the Beran estimators increases. However, one can see also that this growth is
slowing down. It can also be seen from Fig. \ref{fig:estimator_dependency}
that SurvBETA with optimization of parameters outperforms SurvBETA without
optimization. An important result is that SurvBETA significantly outperforms
the single Beran estimator.%

\begin{figure}
[ptb]
\begin{center}
\includegraphics[
height=2.7032in,
width=3.5976in
]%
{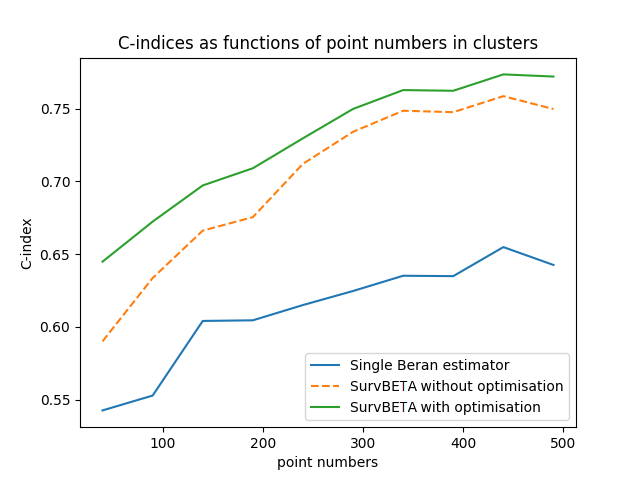}%
\caption{Dependence of the C-index on the number of points in each cluster}%
\label{fig:cluster_points_dependency}%
\end{center}
\end{figure}

\emph{Dependence of the C-index on the number of points in each cluster:} the
number of points varies in the range $[40,500]$ with step $50$. The
corresponding dependence is shown in Fig. \ref{fig:cluster_points_dependency},
where all models illustrate the growth of the C-index with increase of the
number of points in clusters. It follows from the fact that the sample size
increases and the models are better trained. It is interesting to note that
the single Beran estimator also improves its performance with the number of
points in clusters. However, this improvement in the C-index growth is weaker
than the improvement that the models show. Some convergence of the different
modifications of SurvBETA can also be observed with the increase in the number
of points.%

\begin{figure}
[ptb]
\begin{center}
\includegraphics[
height=2.6784in,
width=3.5647in
]%
{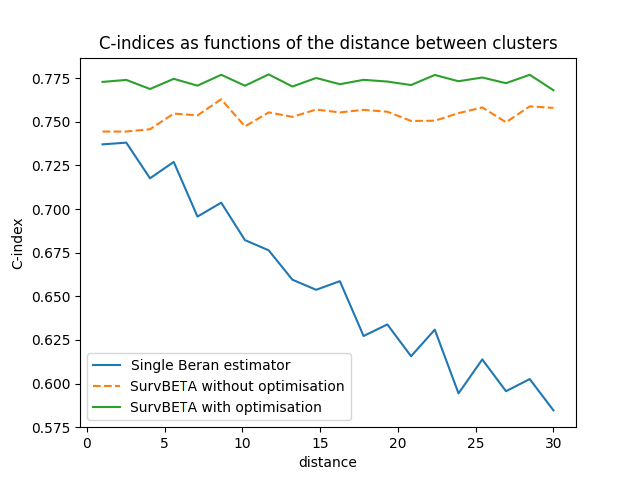}%
\caption{Dependence of the C-index on the distance between cluster}%
\label{fig:cluster_distance_dependency}%
\end{center}
\end{figure}

\emph{Dependence of the C-index on the distance between clusters:} parameters
of the first cluster remains without changes, but hyper-rectangles determining
the second cluster generation are constructed such that their parameters are
shifted relative to the first cluster as follows: $l_{2}^{(j)}=r_{1}^{(j)}+h$,
$r_{2}^{(j)}=l_{2}^{(j)}+10$, $j=1,...,5$, where $h$ varies in the range
$[1,30]$ with step $29/30$. Interesting results are depicted in Fig.
\ref{fig:cluster_distance_dependency}. It can be seen from Fig.
\ref{fig:cluster_distance_dependency} that the single Beran estimator, despite
its kernel basis, does not cope with the situation when the distance between
clusters increases and becomes quite large whereas the proposed models
practically do not deteriorate their performance. This is an important
property of the proposed models which can be regarded as one of the reasons
why the models have been developed.

\begin{figure}
[ptb]
\begin{center}
\includegraphics[
height=2.592in,
width=3.4494in
]%
{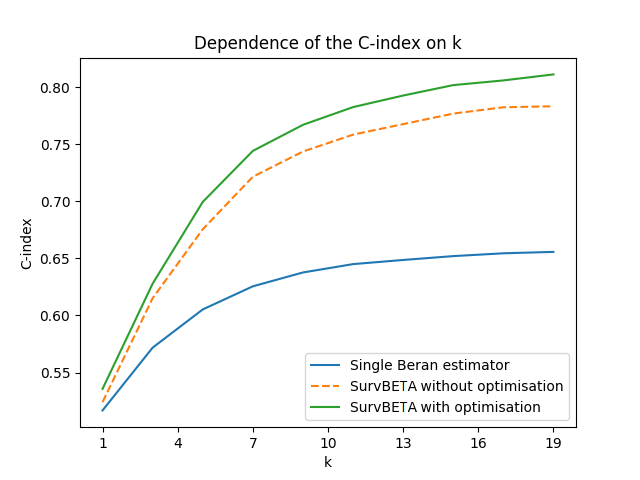}%
\caption{Dependence of the C-index on the parameter $k$}%
\label{fig:k_parameter_dependency}%
\end{center}
\end{figure}

\emph{Dependence of the C-index on the parameter }$k$\emph{:} parameter $k$
varies in the range $[1,21]$ with step $2$. We study how the C-index of the
models depends on the parameter $k$ of the Weibull distribution. In
particular, if $k=1$, then the exponential distribution is used for
generation. If $k=2$, then the Rayleigh distribution is used in this case. As
the parameter increases, the event time is more concentrated around a certain
value, which should reduce the uncertainty of predictions. This is precisely
reflected in Fig. \ref{fig:k_parameter_dependency} depicting the dependence of
the C-index on the parameter $k$ for different models. It can be seen that the
C-index increases with increasing $k$, which fully corresponds to the
intuitive idea of {}{}how the model should behave. At the same time, it can be
seen from Fig. \ref{fig:k_parameter_dependency} that the proposed models
demonstrate better characteristics compared to the single Beran estimator.%

\begin{figure}
[ptb]
\begin{center}
\includegraphics[
height=2.7723in,
width=3.689in
]%
{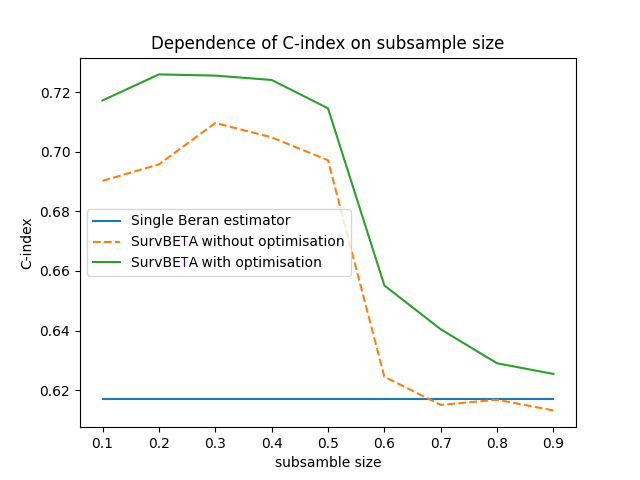}%
\caption{Dependence of the C-index on the subsample size}%
\label{fig:subset_size_dependency}%
\end{center}
\end{figure}

\emph{Dependence of the C-index on the bootstrap subsample size:} the
bootstrap subsample size varies in the range $[0.1,0.9]$ with step $0.1$ (here
the portion of the training set is considered). Fig.
\ref{fig:subset_size_dependency} illustrates how the bootstrap subsample size
impacts on the model performance. It can be seen from Fig.
\ref{fig:subset_size_dependency} that there are some optimal values of the
subsample size, which depend on the model analyzed. In particular, optimal
values of portions are $0.2$ and $0.3$ for SurvBETA with optimization and
without optimization, respectively. It can also be seen that the SurvBETA
performance converges to the single Beran estimator performance when the size
of bootstrap subsamples increases. This is due to the fact that the samples
become too dependent in order to implement a proper ensemble.

\subsection{Real data}

In order to compare SurvBETA with different survival models on real data, we
use the following survival models: the single Beran estimator, SurvBETA
without optimization, SurvBETA with optimization, RSF \cite{Ibrahim-etal-2008}%
, GBM Cox \cite{ridgeway1999state}, GBM AFT \cite{Barnwal-etal-22}.

The models is tested on the following real benchmark datasets.

The \emph{Veterans' Administration Lung Cancer Study (Veteran) Dataset}
\cite{Kalbfleisch-Prentice-1980} contains observations of 137 patients
characterized by 8 features. It can be obtained via the \textquotedblleft
survival\textquotedblright \ R package.

\emph{The AIDS Clinical Trials Group Study (AIDS) Dataset}
\cite{hammer1996trial} contains healthcare statistics and categorical
information about 1151 patients with 19 features who have been diagnosed with
AIDS. It is available at https://www.kaggle.com/datasets/tanshihjen/aids-clinical-trials.

The $($\emph{Breast Cancer) Dataset} \cite{desmedt2007strong} has 198 samples
and 80 features. The endpoint is the presence of distance metastases, which
occurred for 51 patients (25.8\%). The dataset can be obtained via the Python
\textquotedblleft scikit-survival\textquotedblright \ package.

The \emph{German Breast Cancer Study Group 2 (GBSG2) Dataset}\textbf{
}contains observations of 686 patients with 9 features
\cite{Sauerbrei-Royston-1999}. It can be obtained via the \textquotedblleft
TH.data\textquotedblright \ R package.

The \emph{Worcester Heart Attack Study (WHAS500) Dataset}
\cite{Hosmer-Lemeshow-May-2008} considers 500 patients with 14 features. The
dataset can be obtained via the \textquotedblleft smoothHR\textquotedblright%
\ R package or the Python \textquotedblleft scikit-survival\textquotedblright \ package.

The \emph{Primary Biliary Cirrhosis (PBC) Dataset} studies 418 patients with
17 features \cite{Fleming-Harrington-1991}. It has 257 censored observations.
The dataset can be downloaded via the \textquotedblleft
randomForestSRC\textquotedblright \ R package.

The \emph{Wisconsin Prognostic Breast Cancer (WPBC) Dataset}\textbf{
}\cite{Street_etal_95} contains 198 instances having 32 features. It can be
obtained via the \textquotedblleft TH.data\textquotedblright \ R package.

The \emph{Heart Transplant Dataset (HTD)}\textbf{ }consists of observations of
103 patients with 8 features \cite{Kalbfleisch-Prentice-1980}. This dataset is
available at http://lib.stat.cmu.edu/datasets/stanford.

The \emph{Chronic Myelogenous Leukemia Survival (CML) Dataset} is simulated
according to structure of the data by the German CML Study Group used in
\cite{Hehlmann-etal-1994}. The dataset consists of 507 observations with 8
features: a factor with 54 levels indicating the study center; a factor with
levels trt1, trt2, trt3 indicating the treatment group; sex (0 = female, 1 =
male); age in years; risk group (0 = low, 1 = medium, 2 = high); censoring
status (FALSE = censored, TRUE = dead); time survival or censoring time in
days. The dataset can be obtained via the \textquotedblleft
multcomp\textquotedblright \ R package (cml).

The \emph{Molecular Taxonomy of Breast Cancer International Consortium
(METABRIC) Database }\cite{Curtis-etal-12}\emph{ }contains the records of 1904
breast cancer patients and has 9 gene indicator features and 5 clinical features.

The \emph{North Central Cancer Treatment Group (Lung Cancer) Dataset}\textbf{
}\cite{loprinzi1994prospective} contains observations of 228 patients with 7
features and shows performance scores rate how well the patient can perform
usual daily activities. The dataset can be downloaded via the
\textquotedblleft survival\textquotedblright \ R package.

The \emph{(Rossi) Dataset }\textbf{ }\cite{fox2012rcmdrplugin}
contains 432 convicts who were released from Maryland state prisons
in the 1970s. The dataset has 7 features.

SurvBETA is implemented by means of a software in Python. The corresponding
software is available at
\url{https://github.com/NTAILab/SurvBETA}.

The following sets of hyperparameters are used to implement the compared models:

\begin{enumerate}
\item \emph{SurvBETA with optimization}: the learning rate is from
$\{10^{-3},10^{-2},10^{-1}\}$; the number of epochs is taken from the interval
$[50,500]$; the largest number of pairs for each instance is from the interval
$[1,p_{\max}]$; $\epsilon \in \lbrack0,1]$; $w\in \{10^{-3},10^{-2}%
,10^{-1},10^{0},10^{1},10^{2},10^{3}\}$; the bootstrap subsample size varies
in the interval $[0.1,0.7]$; the number of the Beran estimators $M$ is from
the interval $[10,e_{\max}]$.

\item \emph{SurvBETA without optimization}: $w\in \{10^{-4},10^{-3}%
,10^{-2},10^{-1},10^{0},10^{1},10^{2},10^{3},10^{4}\}$; the bootstrap
subsample size varies in the interval $[0.1,0.7]$; the number of the Beran
estimators $M$ is taken from $[10,e_{\max}]$.

\item \emph{The single Beran estimator}: $\tau \in \{10^{-3},10^{-2}%
,10^{-1},10^{0},10^{1},10^{2},10^{3}\}$.

\item \emph{RSF}: the number of trees in the RSF varies in the interval
$[20,500]$; the largest tree depth varies in the interval $[2,15]$; the
smallest number of instances in each leaf is from the interval $[2,15]$.

\item \emph{GBM Cox}: the learning rate is from $\{10^{-3},10^{-2},10^{-1}\}$;
the number of iterations is from the interval $[20,500]$; the largest tree
depth varies in the interval $[2,10]$.

\item \emph{GBM AFT}: the learning rate is from $\{10^{-3},10^{-2},10^{-1}\}$;
the number of iterations $\in \lbrack20,500]$; the largest tree depth varies in
the interval $[2,10]$.
\end{enumerate}

Parameters $e_{\max}$ and $p_{\max}$ depend on the dataset analyzed. They are
equal to the number of instances in the training set.

The comparison results of different models for the above datasets are shown in
Table \ref{results_real_data}. It can be seen from Table
\ref{results_real_data} that SurvBETA with optimization outperforms other
models for most datasets. It is interesting that RSF and GBM Cox also show the
outperforming results for some datasets.%

\begin{table}[tbp] \centering
\caption{C-indices obtained for different datasets by using the single Beran estimator, SurvBETA without optimization, SurvBETA with optimization, RSF, GBM Cox, GBM AFT.}%
\begin{tabular}
[c]{ccccccc}\hline
& Beran & SurvBETA & SurvBETA & RSF & GBM & GBM\\
Datasets & estimator & without opt. & with opt. &  & Cox & AFT\\ \hline
Veterans & $0.6346$ & $0.7051$ & $\mathbf{0.7254}$ & $0.7001$ & $0.6821$ &
$0.7004$\\
AIDS & $0.6466$ & $0.7340$ & $\mathbf{0.7460}$ & $0.7339$ & $0.7198$ &
$0.5800$\\
Breast Cancer & $0.6595$ & $0.7045$ & $\mathbf{0.7184}$ & $0.6806$ & $0.7021$
& $0.6979$\\
GBSG2 & $0.6552$ & $0.6764$ & $0.6826$ & $\mathbf{0.6885}$ & $0.6823$ &
$0.6777$\\
Whas500 & $0.7181$ & $0.7580$ & $\mathbf{0.7644}$ & $0.7610$ & $0.7528$ &
$0.7492$\\
PBC & $0.7619$ & $0.7781$ & $0.7879$ & $0.8099$ & $\mathbf{0.8153}$ &
$0.7969$\\
WPBC & $0.6495$ & $0.7191$ & $\mathbf{0.7499}$ & $0.6387$ & $0.6159$ &
$0.6078$\\
HTD & $0.7421$ & $0.8099$ & $\mathbf{0.8279}$ & $0.7809$ & $0.7749$ &
$0.7932$\\
CML & $0.7009$ & $0.7062$ & $0.7100$ & $0.7129$ & $\mathbf{0.7136}$ &
$0.7114$\\
METABRIC & $0.5879$ & $0.6017$ & $0.6244$ & $\mathbf{0.6334}$ & $0.6301$ &
$0.6321$\\
Lung & $0.6256$ & $0.6611$ & $\mathbf{0.6741}$ & $0.6115$ & $0.5799$ &
$0.5588$\\
Rossi & $0.5245$ & $0.6023$ & $\mathbf{0.6177}$ & $0.5182$ & $0.5117$ &
$0.5380$\\ \hline
\end{tabular}
\label{results_real_data}%
\end{table}%

In order to formally show that SurvBETA with optimization outperforms other
methods, we apply the $t$-test, which has been considered by Demsar
\cite{Demsar-2006} for testing whether the average difference in the
performance of a pair of classifiers is significantly different from zero. The
$t$ statistic in this case is distributed according to the Student's
$t$-distribution with $12-1$ degrees of freedom. The $t$ statistics for the
differences lead to the p-values. The p-values for pairs of models (SurvBETA
with optimization - another model) and (SurvBETA without optimization -
another model) are shown in the corresponding rows of Table \ref{t:t-test1}.
It can be seen from the results provided in Table \ref{t:t-test1} that the
proposed model significantly outperforms all the considered models because the
corresponding p-values are less than $0.05$. One can also see from the results
that SurvBETA without optimization mostly outperforms other models, but this
outperformance cannot be viewed as significant because the corresponding
p-values are not less than $0.05$. At the same time, SurvBETA without
optimization significantly outperforms the single Beran estimator.%

\begin{table}[tbp] \centering
\caption{P-values of the model comprison}%
\begin{tabular}
[c]{cccccc}\hline
& Beran & SurvBETA & RSF & GBM & GBM\\
& estimator & without opt. &  & Cox & AFT\\ \hline
SurvBETA with opt. & $0.00007$ & $0.00007$ & $0.035$ & $0.038$ & $0.019$\\
SurvBETA without opt. & $0.0002$ & $-$ & $0.194$ & $0.112$ & $0.068$\\ \hline
\end{tabular}
\label{t:t-test1}%
\end{table}%

The obtained results of numerical experiments show that the proposed model
SurvBETA with trainable parameters can be regarded as an alternative tool for
survival analysis when datasets have a complex structure.

\section{Conclusion}

A new method called SurvBETA for solving the survival analysis problem under
censored data has been presented. It extends and combines ideas behind the
Beran estimator and ensemble models to provide a more flexible and robust
model. The model is based on three-level attention mechanism. The first
attention level is a part of the Beran estimators which assigns weights to
pairs of instances. The second attention level aims to find prototypes of
bootstrap subsamples related to a new analyzed instance. The first and second
levels can be viewed as local attention mechanisms because the mechanisms are
applied to weak learners of the ensemble whereas the third attention level is
global. It aims to aggregate predictions of all Beran estimators.

The proposed model architecture is flexible. It allows us to replace the Beran
estimators with different models. In this case, the second and third attention
mechanisms are not changed. This is important peculiarity of the model because
it is able to deal with multimodal data. For example, the Beran estimators as
a part of the ensemble model can be trained on tabular data whereas
transformer-based models \cite{hu2021transformer} as another part of the
ensemble models can be trained on images. The whole architecture can also be
trained in the end-to-end manner. Various modifications of the ensemble-based
architectures can be regarded as a direction fur further research.

On the one hand, we have presented simple modifications of the general model
when the imprecise Huber's $\epsilon$-contamination model \cite{Huber81} is
used to implement the third-level attention mechanism. On the other hand, the
general problem has been stated, but not implemented due to the large number
of training parameters and the complexity procedure of training. However,
several approaches can be applied to implementing intermediate modifications
which could provide better results. This is another direction for further research.

Explicit kernels, for example, the Gaussian kernels, have been used in
attention mechanisms of all attention levels. However, an interesting
direction for further research is to apply universal kernels implemented by
means of neural networks which are trained among other training parameters of
the whole model. In this case, the neural networks can significantly extend
the set of possible ensemble-based models.

\bibliographystyle{plain}
\bibliography{Attention,Boosting,Classif_bib,Deep_Forest,MYBIB,MYUSE,Survival_analysis}

\end{document}